\documentclass[10pt,twocolumn,letterpaper]{article}

\usepackage[pagenumbers]{wacv} % To force page numbers, e.g. for an arXiv version

\usepackage{subcaption}
\usepackage{algorithm}
\usepackage{algorithmic}
\usepackage{amsfonts}
\usepackage{amsmath}
\usepackage{booktabs}
\usepackage{caption}
\usepackage{graphicx}
\usepackage{makecell}
\usepackage{multirow}
\usepackage{pifont}
\usepackage{placeins}
\usepackage{xcolor}
\usepackage{color}
\usepackage{graphicx}
\usepackage{newfloat}
\usepackage{listings}
\usepackage[table]{xcolor}
\definecolor{lightblue}{RGB}{224,243,255}  % 很接近你图里的浅蓝
\usepackage{subcaption}
\usepackage{amsmath}
\usepackage{amssymb}

\AtBeginDocument{%
  \setlength{\abovedisplayskip}{6pt plus 1pt minus 1pt}
  \setlength{\belowdisplayskip}{6pt plus 1pt minus 1pt}
  \setlength{\abovedisplayshortskip}{4pt plus 1pt minus 1pt}
  \setlength{\belowdisplayshortskip}{6pt plus 1pt minus 1pt}
  \setlength{\jot}{2pt}
}

\newtheorem{rem}{Remark}

\definecolor{wacvblue}{rgb}{0.21,0.49,0.74}
\usepackage[pagebackref,breaklinks,colorlinks,allcolors=wacvblue]{hyperref}

\def\wacvPaperID{732} % *** Enter the WACV Paper ID here
\def\confName{WACV}
\def\confYear{2027}

\title{SUMO: Segment and Track Any Motion with Nonlinear State Space Models\vspace{-0.3cm}}

\author{
Kexin Tian$^{1}$\thanks{Equal contribution.}\qquad
Sixu Li$^{1}$\footnotemark[1]\qquad
Keshu Wu$^{1}$\qquad
Yang Zhou$^{1}$\thanks{Corresponding author.}\qquad
Zhengzhong Tu$^{1}$\\[0.6em]
$^{1}$Texas A\&M University
}

\begin{document}
\maketitle
\begin{abstract}
Visual Object Tracking (VOT) and Moving Object Segmentation (MOS) are two fundamental tasks in computer vision that involve both spatial and temporal object dynamics. 
    Existing methods rely predominantly on visual cues and thus often falter in real-world scenarios where object motions are inherently complex and nonlinear. 
    To address this limitation, we propose \textbf{SUMO}—a zero-shot, training-free, unified framework integrating nonlinear dynamics with vision-based segmentation for accurate and consistent VOT and MOS. 
    Specifically, we develop a nonlinear State Space Model (SSM) inspired by robotics principles to capture the complex object dynamics. Building on this model, we propose a Selective Unscented Filter (SUF) for accurate state estimation, which features a joint scoring mechanism and dynamically fuses multi-source predictions to identify the most plausible object state over time. 
    Furthermore, we apply a memory selection mechanism to evaluate the reliability of memory frames.
    Our extensive experimental results show that SUMO achieves state-of-the-art performance on \textbf{both} VOT and MOS tasks. The code will be released at \href{https://github.com/kxint9625/SUMO}{\texttt{github.com/kxint9625/SUMO}}.
\end{abstract}

\section{Introduction}
\label{sec:intro}

Visual Object Tracking (VOT)~\cite{muller2018trackingnet,bertinetto2016fully} and Moving Object Segmentation (MOS)~\cite{huang2025segment,bosch2021deep} are two critical and fundamental tasks in computer vision, serving as essential components in a broad range of real-world applications, including autonomous driving~\cite{Ding_2024_CVPR,tian2025nuscenes} and robotics~\cite{8705685,8995968}. 

VOT aims to identify and localize a target object across frames based on its initial status, using a bounding box to represent its position and scale~\cite{xie2024autoregressive}, while MOS aims to precisely segment a moving object by capturing its contour through a pixel-level mask~\cite{huang2025segment}. Despite the differences, both tasks face challenges in maintaining both high accuracy and temporal consistency, particularly under challenging conditions such as complex motion and occlusion.

Recent advances in foundation models have substantially enhanced computer vision tasks. For instance, the Segment Anything Model (SAM)~\cite{kirillov2023segment} achieves high-precision segmentation masks across diverse visual scenarios. Building upon SAM, the more recent SAM2~\cite{ravi2024sam} further extends this capability to video sequences, enabling frame-by-frame segmentation of objects in videos.
\begin{figure}[t]
    \centering
{\includegraphics[width=0.23\textwidth]{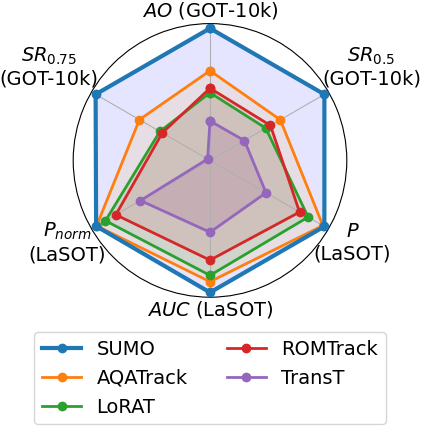}}
{\includegraphics[width=0.23\textwidth]{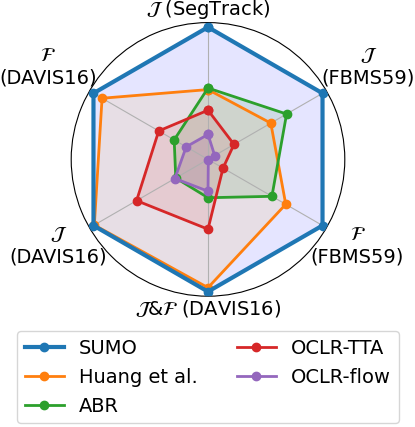}}
\vspace{-0.2cm}
    \caption{Comprehensive experiments demonstrate that SUMO achieves state-of-the-art performance on both VOT (left) and MOS (right) tasks. The radar plots visualize the normalized performance across multiple benchmarks and evaluation metrics.}
    \label{fig: reparameterization}
    \vspace{-0.5cm}
\end{figure}

However, SAM2 remains limited in its performance on VOT and MOS, as it relies solely on visual appearance without modeling underlying dynamics. In real-world scenarios, however, objects often exhibit complex dynamics~\cite{tian2025physically}, accompanied by frequent appearance deformation and occlusions, leading to significant visual variation over time. As a result, purely visual cues are often insufficient for maintaining object identity across frames, especially in VOT and MOS that require consistent spatiotemporal association. These limitations underscore the need to model object dynamics as a complement to visual appearance.

Object dynamics in time-series data can be modeled using State Space Models (SSMs), which offer a mathematical framework for capturing the temporal evolution of underlying system states~\cite{callier2012linear}. Recent work such as Mamba~\cite{gu2023mamba} demonstrates that integrating SSMs into relatively simple neural architectures achieve performance comparable to transformers with greater computational efficiency. Subsequent studies further validate SSMs for modeling long-range dependencies in diverse domains~\cite{liu2024vmamba,ma2024u,liang2024pointmamba}.
More recently, SAMURAI~\cite{yang2024samurai} augments the SAM2 with a Kalman Filter to improve temporal consistency in VOT. 

However, the SSM in Mamba and the Kalman filter in SAMURAI are both limited to linear systems.
In practice, most systems are inherently nonlinear, where adopting linear approaches imposes strong inductive biases. 
Therefore, it is essential to employ nonlinear approaches for accurate dynamics modeling~\cite{li2025nonlinear}, especially in VOT and MOS tasks, where better capturing the nonlinear object dynamics can lead to more accurate and robust performance.

To address this gap, we introduce \textbf{SUMO}—\underline{\textbf{S}}egment and track any motion \underline{\textbf{U}}sing nonlinear state space \underline{\textbf{MO}}dels—a zero-shot, training-free, unified framework that integrates nonlinear dynamics with vision-based segmentation to achieve accurate and temporally consistent VOT and MOS.
We first develop a nonlinear SSM to capture the object dynamics. To enable accurate state estimation in VOT and MOS, where explicit sensor measurements are unavailable, we propose a Selective Unscented Filter (SUF) that integrates with a purely vision-based video segmentation framework, SAM2, in a plug-and-play manner. The SUF combines predictions from both the nonlinear SSM and a mask decoder, and applies a joint scoring mechanism to select the most plausible mask. The selected mask is then treated as the measurement input in the SUF's update step, allowing the filter to refine the SSM state estimate for the next video frame. Moreover, a memory selection mechanism is applied to evaluate the reliability of memory frames.

Our main contributions are as follows:

\begin{itemize}
    \item We present a unified framework that integrates nonlinear dynamics with vision-based segmentation, enabling accuracy and temporal consistency in both VOT and MOS.
    \item We design a nonlinear SSM, inspired by robotics principles, to capture complex and realistic object dynamics.
    \item We propose a SUF that enables accurate state estimation in the absence of explicit sensor measurements, integrating seamlessly with SAM2 in a plug-and-play manner.
    \item SUMO achieves the state-of-the-art performance in both VOT and MOS tasks.
\end{itemize}
\section{Related Works}
\label{sec:related_works}

\subsection{VOT and MOS}

Traditionally, \textbf{VOT} methods are broadly categorized into three paradigms. Siamese-based trackers perform tracking by comparing a fixed template from the initial frame with the current search region using similarity matching~\cite{bertinetto2016fully,li2018high,li2019siamrpn++,chen2020siamese}. Discriminative trackers learn a target-specific classifier online to distinguish the object from the background and update it during tracking~\cite{zhang2017mdnet,bhat2019learning,danelljan2019atom}. While Siamese-based methods often struggle with large appearance variations, discriminative approaches rely on expensive online updates. To address these limitations, transformer-based trackers have emerged, using attention mechanisms to model rich spatial and temporal dependencies for robust feature fusion~\cite{chen2021transformer,mayer2022transforming,cui2022mixformer,zhang2019deeper}.
Meanwhile, existing \textbf{MOS} methods can be broadly categorized into flow-based approaches~\cite{xie2024moving,bosch2021deep,bideau2016s,cao2019learning}, which rely on optical flow to extract short-term pixel-wise motion, and trajectory-based approaches~\cite{karazija2024learning, lai2016motion,barath2019progressive}, which cluster point trajectories to identify groups exhibiting consistent motion and capture long-range motion patterns.
However, most existing methods in both tasks are purely vision-based and struggle to capture the complex, nonlinear dynamics of objects in real-world scenarios, particularly under challenging conditions such as occlusion, shape deformation, and abrupt movement.

\subsection{SSMs}

% Recent studies have increasingly integrated SSMs into machine learning. Mamba~\cite{gu2023mamba} leverages input-conditioned SSMs for efficient long-range modeling, outperforming transformers on diverse sequence tasks with lower computational cost, demonstrating the strength of SSMs in neural architectures.
% A series of follow-up works have extended SSM integration to various domains. For example, VMamba~\cite{liu2024vmamba} adapts SSMs to visual tasks, achieving strong results in image classification and dense prediction. U-Mamba~\cite{ma2024u} and SegMamba~\cite{xing2024segmamba} apply SSMs to biomedical segmentation, enhancing the long-range dependencies modeling in medical images. PointMamba~\cite{liang2024pointmamba} incorporates SSMs into point cloud analysis. These works collectively demonstrate the strong performance and versatility of SSMs across diverse domains.
% However, all of these methods are built upon linear SSMs, which can only capture linear dynamics or rely on linear assumptions. In practice, real-world object dynamics is often complex and highly nonlinear. Therefore, relying solely on linear SSMs is inadequate for VOT and MOS

Recent studies have increasingly integrated SSMs into modern machine learning architectures. Mamba~\cite{gu2023mamba} leverages input-conditioned SSMs for efficient long-range modeling, outperforming transformers on a variety of sequence modeling tasks with lower computational cost, thereby demonstrating the effectiveness of SSMs in neural architectures.
A series of follow-up works have further extended the integration of SSMs to a wide range of domains. For example, VMamba~\cite{liu2024vmamba} adapts SSMs to visual tasks, achieving strong performance in image classification and dense prediction. U-Mamba~\cite{ma2024u} and SegMamba~\cite{xing2024segmamba} apply SSMs to biomedical segmentation, enhancing the modeling of long-range dependencies in medical images. PointMamba~\cite{liang2024pointmamba} incorporates SSMs into point cloud analysis for 3D understanding. These works collectively demonstrate the strong performance and versatility of SSM-based architectures across diverse domains.
However, all of these methods are built upon linear SSMs, which can only capture linear dynamics or rely on simplified linear assumptions. In practice, real-world object dynamics are often complex and highly nonlinear. Therefore, relying solely on linear SSMs is inadequate for effectively modeling the complex real-world dynamics required in VOT and MOS tasks.
\section{Methodology}
\label{sec:methodology}

\begin{figure*}[!t]
    \centering
    \includegraphics[width=0.9\linewidth]{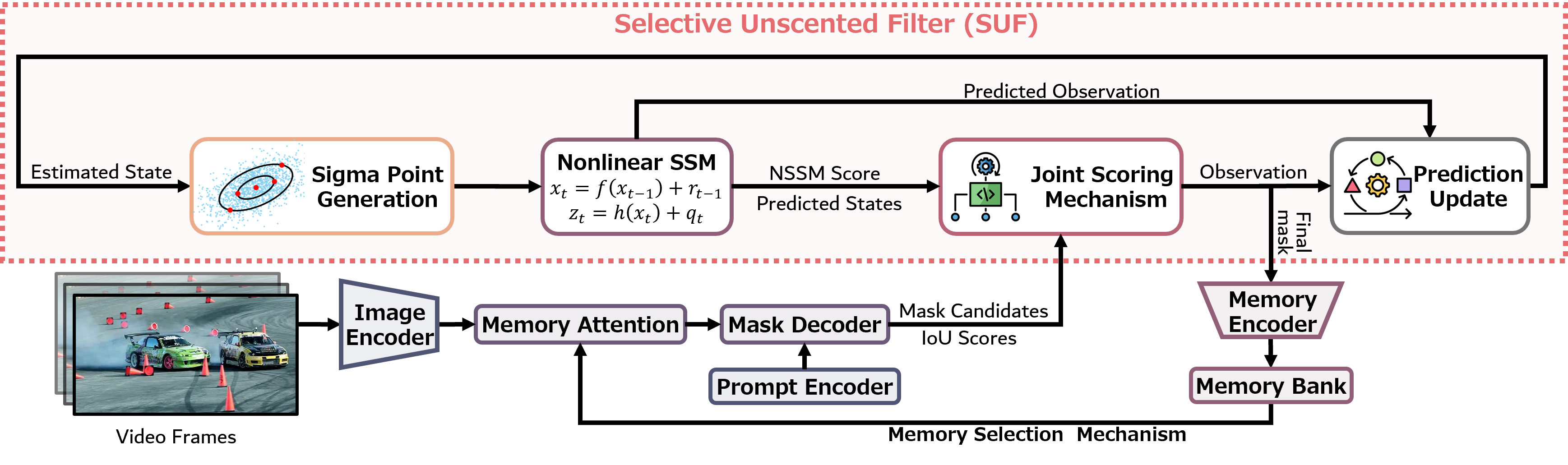}
    \vspace{-0.1cm}
    \caption{Framework Architecture.}
    \label{fig: framework}
    \vspace{-0cm}
\end{figure*}

This section presents the methodology of the SUMO framework. We first describe the vision modules in~\cref{sec:vision_modules}, followed by the nonlinear State Space Model in~\cref{sec:nonlinearssms}, the Selective Unscented Filter in~\cref{sec:suf}, and the Memory Selection Mechanism in~\cref{sec:mmm}. An overview of the framework is shown in~\cref{fig: framework}.

\subsection{Vision Modules}
\label{sec:vision_modules}

\subsubsection{Image Encoder} 

% The image encoder extracts multi-scale image features once per frame, producing hierarchical feature maps that serve as the foundation for both memory construction and subsequent decoding modules. For the image encoder module, We retain the architecture from SAM2~\cite{ravi2024sam}, which employs a Hiera~\cite{ryali2023hiera} model pretrained with masked autoencoding~\cite{he2022masked}.

The image encoder extracts multi-scale features once per frame, producing hierarchical feature maps for memory construction and subsequent decoding. We adopt the SAM2~\cite{ravi2024sam} encoder architecture, which uses a Hiera~\cite{ryali2023hiera} model pretrained with masked autoencoding~\cite{he2022masked}.

\subsubsection{Memory Attention}

This module integrates selected memory frames into the current frame via a memory selection mechanism, which will be detailed in Eq.\eqref{eq:memory_selection}, and a transformer-based attention architecture that follows the SAM2\cite{ravi2024sam} design, which applies self-attention to the current frame features followed by cross-attention to the memory features.

\subsubsection{Prompt Encoder}

In the initial frame, a bounding box specifying the tracking target is provided as the prompt. The prompt encoder processes this input, which is then passed to the mask decoder to generate the initial object mask for tracking.

\subsubsection{Mask Decoder}

The mask decoder takes as input the fused features from the memory attention module and high-resolution image features from the image encoder. It then decodes three mask candidates, each with an IoU score, an objectness score indicating the object's presence, and an object pointer, which is a compact embedding encoding the object's identity and appearance. All three mask candidates are retained for the subsequent  SUF process.

\subsection{Nonlinear SSMs}
\label{sec:nonlinearssms}

\subsubsection{Brief Review of Linear SSM}

Linear SSM (or more precisely, state-space \textit{representation} of linear systems \cite{callier2012linear}) provides a principled framework for representing the evolution of linear dynamical systems. In the discrete-time setting, the system state $\mathbf{x}_t \in \mathbb{R}^n$ at time step $t$ is determined by its previous state $\mathbf{x}_{t-1}$ and a control input $\mathbf{u}_{t-1} \in \mathbb{R}^m$. The dynamics are typically described by the following first-order difference equation:
\begin{equation}
    \mathbf{x}_t = \mathbf{A} \mathbf{x}_{t-1} + \mathbf{B} \mathbf{u}_{t-1},
\end{equation}
where $\mathbf{A} \in \mathbb{R}^{n \times n}$ is the state matrix that characterizes the intrinsic dynamics of the system, and $\mathbf{B} \in \mathbb{R}^{n \times m}$ encodes how the control input influences the state evolution. In reality, the dynamics of nearly all systems are nonlinear. Representing them with linear SSMs generally entails linearization or model simplification, possibly resulting in notable approximation errors and restricted representational ability.

\subsubsection{Nonlinear SSM}    
\label{sec:nonlinear-ssm}
Nonlinear SSM is a generalization of linear SSM and offers greater expressive power and improved accuracy in modeling real-world dynamics \cite{khalil2002nonlinear}. For a discrete-time system, the dynamics are described as
\begin{equation}
\label{eq: nSSM_general}
    \mathbf{x}_t = f(\mathbf{x}_{t-1}, \mathbf{u}_{t-1}),
\end{equation}
where $f(\cdot)$ is a vector-valued (nonlinear) function.

In robotics \cite{lynch2017modern}, a simple yet versatile way to describe planar motion is to incorporate the notion of a heading direction  \cite{dubins1957curves,sussmann1991shortest}. Such models can represent various systems, including wheeled robots, ground vehicles, airplanes, bicycles, rolling robots, and even humans or animals (e.g., when walking or running in the direction they face) \cite{ben2021time,ibrahim2025accelerated,li2025time,li2025closed,kumar2023weighted,li2024sequencing}. To accommodate more agile motions, we extend the model in \cite{sussmann1991shortest} by allowing the heading direction to be instantaneously controlled and apply the resulting formulation to the image plane. Additionally, we augment the state with variables describing the object’s bounding box.

\paragraph{State}

For our nonlinear SSM, we define the state vector $\mathbf{x} \in \mathbb{R}^8$ as
\begin{equation}
\label{eq:state_vector}
\mathbf{x} = [x, y, a, h, v, \theta, v_a, v_h]^T,
\end{equation}
where $x$ and $y$ denote the center of the bounding box in pixel coordinates. $h$ represents the height of the bounding box and $a$ the aspect ratio of the bounding box ($a = w/h$). $v$ and $\theta$ denote the speed and heading direction of motion, respectively. $v_a$ and $v_h$ represent the rate of change of $a$ and $h$, respectively. This formulation jointly captures the object dynamics and shape variations. 
\begin{rem}
Note that if the sensor measurements of $v, \theta, v_a, v_h$ were available, they could be grouped as the control input $\mathbf{u}$ in Eq.~\eqref{eq: nSSM_general}. However, such data are unavailable in VOT and MOS. Hence, we include these variables in the state vector for estimation and set $\mathbf{u}_t$ as zero for all $t$. 
\end{rem}

\paragraph{Nonlinear State Equation}

To represent the temporal evolution of the state, we introduce a nonlinear state equation $f$ that predicts the state $\mathbf{x}_t$ based on $\mathbf{x}_{t-1}$. Additionally, we introduce a process noise $\mathbf{r}_{t-1}$ for unknown effects:
\begin{equation}
\label{eq:nssm}
\mathbf{x}_t = f(\mathbf{x}_{t-1})+\mathbf{r}_{t-1},
\end{equation}
Let $f_i$ denote the $i$th row of $f(\cdot)$, we have
\begin{align}
    &f_1=x_{t-1} + v_{t-1} \cdot \cos(\theta_{t-1}) \cdot \Delta t,  \\
    &f_2=y_{t-1} + v_{t-1} \cdot \sin(\theta_{t-1}) \cdot \Delta t,  \\
    &f_3=a_{t-1} + v_{a,t-1} \cdot \Delta t, ~~ f_4= h_{t-1} + v_{h,t-1} \cdot \Delta t,  \\
    &f_5=v_{t-1},~~f_6=\theta_{t-1}, ~~f_7=v_{a,t-1}, ~~f_8=v_{h,t-1},
\end{align}

where $\Delta t$ denotes the time interval between two frames.

This state equation introduces nonlinearity through the trigonometric dependency on $\theta$ \cite{sussmann1991shortest}, allowing the system to naturally model curved motion trajectories in the image plane. In addition, it explicitly accounts for temporal changes in the target's shape via $v_a$ and $v_h$. 

\paragraph{Observation Model}

We define the observation vector as:
\begin{equation}
\mathbf{z}_t = [x_t, y_t, a_t, h_t]^\top,
\end{equation}
which is obtained from the selected mask generated by mask decoder. The observation model maps $\mathbf{x_t}$ to $\mathbf{z_t}$:
\begin{equation}
\mathbf{z}_t = h(\mathbf{x}_t)+\mathbf{q}_{t} = \mathbf{H} \mathbf{x}_t+\mathbf{q}_{t},
\end{equation}
where $\mathbf{q}_{t}$ represents the observation noise and 
\begin{equation}
\mathbf{H} = [\mathbf{I}_4 \;\; \mathbf{0}_{4 \times 4}].
\end{equation}

\subsection{Selective Unscented Filter}
\label{sec:suf}
Accurate state estimation under nonlinear dynamics is crucial in real-world scenarios for recovering the state value of the target object from noisy observations. Especially in VOT and MOS, the target object often suffers from occlusions, blurriness, and deformations, making purely appearance-based decisions unreliable. Therefore, it is necessary to estimate the target object’s state throughout the time sequence. 

The Unscented Filter (UF)~\cite{julier2004unscented} is a powerful state estimation method that deals with nonlinear SSMs. It offers significantly enhanced accuracy compared to linearization-based methods, while being derivative-free and well-suited for strongly nonlinear dynamics.
In conventional state estimation tasks, UF relies on measurements from physical sensors to update the system state. However, in VOT and MOS, such measurements are unavailable.

To address this challenge, we propose a \textbf{Selective Unscented Filter} (SUF), built upon the nonlinear SSM introduced in Eq.~\eqref{eq:nssm}, which unifies joint scoring, mask selection, and state estimation into a single framework.

\subsubsection{Sigma Point Generation}

Since a nonlinear function cannot be directly applied to a probability distribution, and computing the output distribution of $f(x)$ is often intractable, the SUF employs the Unscented Transform~\cite{julier2004unscented} for approximation. It does so by deterministically generating a set of sigma points from the current state distribution, which are designed to match its mean and covariance up to second-order accuracy. This enables the uncertainty to be effectively propagated through nonlinear dynamics.

At each timestep $t-1$, given a state vector of dimension $n$, the SUF generates a set of $2n + 1$ sigma points $\{\boldsymbol{\chi}^{(i)}_{t-1}\}_{i=0}^{2n}$ from the current mean $\boldsymbol{\mu}_{t-1} \in \mathbb{R}^n$ and covariance $\boldsymbol{\Sigma}_{t-1} \in \mathbb{R}^{n \times n}$. Each sigma point is assigned with weights $w_i^{(m)}$ and $w_i^{(c)}$ for the computation of the predicted mean and covariance, respectively. We have
\begin{equation}
\boldsymbol{\chi}^{(0)}_{t-1} = \boldsymbol{\mu}_{t-1},
\end{equation}
\begin{equation}
\label{eq: X^i}
\boldsymbol{\chi}^{(i)}_{t-1} = \boldsymbol{\mu}_{t-1} + \sqrt{n + \lambda} \cdot \mathbf{L}_{:, i}, \quad i = 1, \dots, n,
\end{equation}
\begin{equation}
\label{eq: X^i+n}
\boldsymbol{\chi}^{(i+n)}_{t-1} = \boldsymbol{\mu}_{t-1} - \sqrt{n + \lambda} \cdot \mathbf{L}_{:, i}, \quad i = 1, \dots, n.
\end{equation}
\begin{equation}
w_0^{(m)} = \tfrac{\lambda}{n + \lambda}, \quad
w_0^{(c)} = \tfrac{\lambda}{n + \lambda} + (1 - \alpha^2 + \beta),
\end{equation}
\begin{equation}
w_i^{(m)} = w_i^{(c)} = \tfrac{1}{2(n + \lambda)}, \quad i = 1, \dots, 2n.
\end{equation}

The parameters $\alpha$, $\kappa$, and $\beta$ are hyperparameters that determine the spread and overall shape of the sigma point distribution. Here, $\mathbf{L} = \text{Cholesky}(\boldsymbol{\Sigma}_{t-1})$ in Eq.~\eqref{eq: X^i}-\eqref{eq: X^i+n} denotes the lower triangular matrix obtained from the Cholesky decomposition of the covariance matrix $\boldsymbol{\Sigma}_{t-1}$, and $\lambda = \alpha^2 (n + \kappa) - n$ is a scaling factor \cite{wan2000unscented}.

\paragraph{Prediction Step}

Once the sigma points are generated, each sigma point $\boldsymbol{\chi}_{t-1}^{(i)}$ is propagated through the nonlinear state equation $ f(\cdot) $, as defined in Eq.~\eqref{eq:nssm}, yielding:
\begin{equation}
\boldsymbol{\chi}_t^{(i)} = f(\boldsymbol{\chi}_{t-1}^{(i)}).
\end{equation}

Then the mean $\hat{\boldsymbol{\mu}}_x^t$ and covariance $\hat{\boldsymbol{\Sigma}}_x^t$ of the predicted state $\boldsymbol{\hat{x}}_t\sim \mathcal{N}(\hat{\boldsymbol{\mu}}_x^t, \hat{\boldsymbol{\Sigma}}_x^t)$  at time step $t$ are computed as:
\begin{equation}
\hat{\boldsymbol{\mu}}_x^t = \sum_{i=0}^{2n} w_i^{(m)} \boldsymbol{\chi}_t^{(i)},
\end{equation}
\begin{equation}
\hat{\boldsymbol{\Sigma}}_x^t = \sum_{i=0}^{2n} w_i^{(c)} \left( \boldsymbol{\chi}_t^{(i)} - \hat{\boldsymbol{\mu}}_x^t \right) \left( \boldsymbol{\chi}_t^{(i)} - \hat{\boldsymbol{\mu}}_x^t \right)^\top + \Sigma_{rr},
\end{equation}
where $\Sigma_{rr} \in \mathbb{R}^{n \times n}$ is a diagonal process noise covariance matrix, built from predefined standard deviations of the components in the state vector $\mathbf{x}$.

The predicted sigma points are mapped into the observation space using the observation model $h(\cdot)$. In our case, the observation is a bounding box represented as $[x, y, a, h]$, which is a linear subset of the full state vector. Each sigma point is transformed as:
\begin{equation}
\mathbf{\hat{z}}_t^{(i)} = h(\boldsymbol{\chi}_t^{(i)}) = \mathbf{H}\boldsymbol{\chi}_t^{(i)}.
\end{equation}

The predicted observation mean $\hat{\boldsymbol{\mu}}_z^t$ is then computed as:
\begin{equation}
\label{eq:pred_mean}
\hat{\boldsymbol{\mu}}_z^t = \sum_{i=0}^{2n} w_i^{(m)} \mathbf{\hat{z}}_t^{(i)}.
\end{equation}

\paragraph{Joint Scoring Mechanism}

Let $\mathcal{M} = \{m_1, m_2, m_3\}$ denote the set of three mask candidates generated by the mask decoder, each accompanied by a mask decoder score $\text{IoU}_{\text{md}} \in \mathbb{R}^3$ predicted by its IoU head. 

For each mask candidate $m_i \in \mathcal{M}$, we retrieve its tight bounding box $b_i$. Separately, from the predicted observation mean $\hat{\boldsymbol{\mu}}_z^t$ obtained in Eq.~\eqref{eq:pred_mean}, we extract the center coordinates $(x, y)$, height $h$, and aspect ratio $a$, which define the NSSM-predicted bounding box $\hat{b}^{\text{NSSM}}$. We then compute the IoU between $b_i$ and $\hat{b}^{\text{nssm}}$ to obtain the NSSM score for each mask, denoted as $\text{IoU}_{\text{nssm}, i}$.

Finally, each mask candidate is assigned a joint score combining the the NSSM score and the mask decoder score:
\begin{equation}
\text{Score}_i = \lambda \cdot \text{IoU}_{\text{nssm}, i} + (1 - \lambda) \cdot\text{IoU}_{\text{md}, i},
\end{equation}
where $\lambda \in [0, 1]$.
In cases where all $\text{IoU}_{\text{md}, i} < 0.5$, we interpret this as a sign of low decoder confidence, typically due to occlusion, blur, or deformation. In such cases, we fully rely on the bounding box predicted by SUF to guide selection by discarding $\text{IoU}_{\text{md}}$, and instead compute:
\begin{equation}
\text{Score}_i = \text{IoU}_{\text{nssm}, i}.
\end{equation}
% \paragraph{Mask Selection and NSSM Update.}
We select the mask $m_{i^*}$ with the highest final score:
\begin{equation}
i^* = \arg\max_i \text{Score}_i,
\end{equation}
which serves as the final output for the current frame and is further utilized in the downstream update step.

\paragraph{Update Step}

After selecting the final mask, to prevent degradation of the state estimate due to unreliable observations, we introduce a mechanism to determine whether to perform the SUF update step. 

Specifically, we apply a stability threshold $\tau_{s}$ on appearance confidence. If the selected candidate’s appearance IoU does not satisfy $\text{IoU}_{\text{md}, i^*} > \tau_{s}$, the update step is skipped, and we directly use the predicted state as the estimate.

Conversely, if $\text{IoU}_{\text{md}, i^*} > \tau_{s}$, we consider the mask reliable and use it to produce an observation, $\mathbf{z}_t$, and proceed with the SUF update step. 
Specifically, we first compute the predicted observation covariance $\mathbf{\Sigma}_{zz}^t$ at time step $t$ as:
\begin{equation}
\mathbf{\Sigma}_{zz}^t = \sum_{i=0}^{2n} w_i^{(c)} \left(\mathbf{\hat{z}}_t^{(i)} -\hat{\boldsymbol{\mu}}_z^t\right)\left(\mathbf{\hat{z}}_t^{(i)} - \hat{\boldsymbol{\mu}}_z^t\right)^\top + \Sigma_{qq},
\end{equation}
where $\Sigma_{qq} \in \mathbb{R}^{m \times m}$ is a diagonal observation noise covariance matrix, defined from assumed noise levels in the observation’s position and scale.

The cross-covariance $\mathbf{\Sigma}_{\hat{x}z}^t$ between the predicted state and the predicted observation at time step $t$ is computed as:
\begin{equation}
\mathbf{\Sigma}_{\hat{x}z}^t = \sum_{i=0}^{2n} w_i^{(c)} \left(\boldsymbol{\chi}_t^{(i)} - \hat{\boldsymbol{\mu}}_x^t\right)\left(\mathbf{\hat{z}}_t^{(i)} - \hat{\boldsymbol{\mu}}_z^t\right)^\top.
\end{equation}

Then, the update gain is obtained by:
\begin{equation}
\mathbf{K}_t = \mathbf{\Sigma}_{\hat{x}z}^t \left( \mathbf{\Sigma}_{zz}^t \right)^{-1}.
\end{equation}

Finally, the posterior distribution of the estimated state $\boldsymbol{x}_t \sim \mathcal{N}(\boldsymbol{\mu}_t, \boldsymbol{\Sigma}_t)$ at time step $t$ is updated by computing its mean $\boldsymbol{\mu}_t$ and covariance $\boldsymbol{\Sigma}_t$ as:

\begin{equation}
\boldsymbol{\mu}_t = \hat{\boldsymbol{\mu}}_x^t + \mathbf{K}_t \left(\mathbf{{z}}_t - \hat{\boldsymbol{\mu}}_z^t\right),~~
% \end{equation}
% \begin{equation}
\boldsymbol{\Sigma}_t = \hat{\boldsymbol{\Sigma}}_x^t - \mathbf{K}_t \mathbf{\Sigma}_{zz}^t \mathbf{K}_t^\top.
\end{equation}

\subsection{Memory Modules and Mechanism}
\label{sec:mmm}

\paragraph{Memory Encoder} 
fuses high-resolution image features with the final mask to capture appearance and location. It uses a learned no-object embedding when absent and positional encodings for temporal modeling.

\paragraph{Memory Bank}
stores recent contextual features in FIFO queues for memory attention. Each entry contains spatial features, positional encodings, and object pointers for retrieving relevant information over time.

\subsubsection{Memory Selection Mechanism}
\label{sec:memory-selection}

Attending to low-quality memory frames can degrade the prediction of the current frame. Therefore, we applied a memory selection mechanism that selects only reliable frames from the memory bank. A past frame at timestep $t$ is selected for memory attention only if all the following conditions are satisfied: 
\begin{equation}
\label{eq:memory_selection}
\text{IoU}_{\text{md}}^t > \tau_{\text{md}}, \quad
s_{\text{obj}}^t > \tau_{\text{obj}}, \quad
\text{IoU}_{\text{nssm}}^t > \tau_{\text{nssm}}
\end{equation}
Here, for the frame at the timestep $t$, $\text{IoU}_{\text{md}}^t$, $s_{\text{obj}}^t$ and $\text{IoU}_{\text{nssm}}^t$ denote the final mask decoder score, occlusion score, and NSSM score, respectively, with the corresponding thresholds $\tau_{\text{md}}, \tau_{\text{obj}}, \tau_{\text{nssm}}$.

\section{Experiments}
\label{sec:experiments}

\begin{table*}[t]
  \centering
  \resizebox{0.9\textwidth}{!}{
  \begin{tabular}{llccccccccc}
    \specialrule{1.2pt}{0pt}{0pt}
    \toprule
    \multirow{2}{*}{\textbf{Methods}} & \multirow{2}{*}{\textbf{Source}} &
    \multicolumn{3}{c}{\textbf{GOT-10k}} & \multicolumn{3}{c}{\textbf{LaSOT}} & \multicolumn{3}{c}{\textbf{LaSOT-ext}} \\
    \cmidrule(lr){3-5} \cmidrule(lr){6-8} \cmidrule(lr){9-11}
    & & \makecell{AO(\%)} & \makecell{SR$_{0.5}$(\%)} & \makecell{SR$_{0.75}$(\%)} 
      & \makecell{AUC(\%)} & \makecell{P$_{\text{norm}}$(\%)} & \makecell{P(\%)} 
      & \makecell{AUC(\%)} & \makecell{P$_{\text{norm}}$(\%)} & \makecell{P(\%)} \\
    \midrule
    SiamRPN++~\cite{li2019siamrpn++} & CVPR'19 & 51.7 & 61.6 & 32.5 & 49.6 & 56.9 & 49.1 & 34.0 & 41.6 & 39.6 \\
    $\text{TransT}_{256}$~\cite{chen2021transformer} & CVPR'21 & 67.1 & 76.8 & 60.9 & 64.9 & 73.8 & 69.0 & - & - & - \\
    $\text{STARK}_{320}$~\cite{yan2021learning} & ICCV'21 & 68.8 & 78.1 & 64.1 & 67.1 & 76.9 & 72.2 & - & - & - \\
    $\text{SwinTrack-B}_{384}$~\cite{lin2022swintrack} & NeurIPS'22 & 72.4 & 80.5 & 67.8 & 71.4 & 79.4 & 76.5 & - & - & - \\
    $\text{MixFormer}_{288}$~\cite{cui2022mixformer} & CVPR'22 & 70.7 & 80.0 & 67.8 & 69.2 & 78.7 & 74.7 & - & - & - \\
    $\text{OSTrack}_{384}$~\cite{ye2022joint} & ECCV'22 & 73.7 & 83.2 & 70.8 & 71.1 & 81.1 & 77.6 & 50.5 & 61.3 & 57.6 \\
    % $\text{ARTrack-B}_{256}$~\cite{wei2023autoregressive} & ?CVPR'23 & 73.5 & 82.2 & 70.9 & 70.8 & 79.5 & 76.2 & 48.4 & 57.7 & 53.7 \\
    $\text{SeqTrack-B}_{384}$~\cite{chen2023seqtrack} & CVPR'23 & 74.5 & 84.3 & 71.4 & 71.5 & 81.1 & 77.8 & 50.5 & 61.6 & 57.5 \\
    $\text{ROMTrack-B}_{256}$~\cite{cai2023robust} & ICCV'23 & 72.9 & 82.9 & 70.2 & 69.3 & 78.8 & 75.6 & 47.2 & 53.5 & 52.9 \\
    $\text{EVPTrack-B}_{384}$~\cite{shi2024explicit} & AAAI'24 & 76.6 & 86.7 & 73.9 & 72.7 & 82.9 & 80.3 & 53.7 & 65.5 & 61.9 \\
    $\text{ODTrack-B}_{384}$~\cite{zheng2024odtrack} & AAAI'24 & 77.0 & 87.9 & 75.1 & 73.2 & 83.2 & 80.6 & 52.4 & 63.9 & 60.1 \\
    $\text{AQATrack-L}_{384}$~\cite{xie2024autoregressive} & CVPR'24 & 76.0 & 85.2 & 74.9 & 72.7 & 82.9 & 80.2 & 52.7 & 64.2 & 60.8 \\
    $\text{LoRAT-B}_{224}$~\cite{lin2024tracking} & ECCV'24 & 72.1 & 81.8 & 70.7 & 71.7 & 80.9 & 77.3 & 50.3 & 61.6 & 57.1 \\
    $\text{LoRAT-L}_{224}$~\cite{lin2024tracking} & ECCV'24 & 75.7 & 84.9 & 75.0 & 74.2 & \textbf{83.6} & \textbf{80.9} & 52.8 & 64.7 & 60.0 \\
    SAMURAI-L~\cite{yang2024samurai} & ArXiv'25 & 82.7 & 94.5 & 83.4 & 70.1 & 78.1 & 75.5 & 59.0 & 71.7 & 70.1 \\
    \midrule
    % SUMO-T & Ours & 82.0 & 92.9 & 80.1 & 70.5 & 77.5 & 74.8 & 53.5 & 61.4 & 63.4 \\ 
    SUMO-S & Ours & 82.5 & 93.2 & 81.2 & 70.9 & 78.3 & 76.1 & 57.3 & 66.8 & 68.5 \\
    SUMO-B & Ours & 82.0 & 93.0 & 80.2 & 72.8 & 81.0 & 78.6 & 56.7 & 65.8 & 68.2 \\
    SUMO-L & Ours & \textbf{83.5} & \textbf{95.5} & \textbf{83.9} & \textbf{74.4} & 82.8 & 80.5 & \textbf{61.2} & \textbf{72.2} & \textbf{73.9} \\
    \bottomrule
  \end{tabular}
  }
  \vspace{-0.1cm}
  \caption{Comparison of VOT performance across GOT-10k~\cite{huang2019got}, LaSOT~\cite{fan2021lasot}, and LaSOT-ext~\cite{fan2021lasot} datasets.}
  \label{tab:tracker_performance}
  \vspace{0cm}
\end{table*}

\begin{table*}
  \centering{
  \scriptsize
  \resizebox{0.9\linewidth}{!}{
  \begin{tabular}{llccccccccccccc}
    \specialrule{1.2pt}{0pt}{0pt}
    \toprule
    Methods & \textbf{ARC} & \textbf{BC} & \textbf{CM} & \textbf{DEF} & \textbf{FM} & \textbf{FOC} & \textbf{IV} & \textbf{LR} & \textbf{MB} & \textbf{OV} & \textbf{POC} & \textbf{ROT} & \textbf{SV} & \textbf{VC} \\
    \midrule
    TransT~\cite{chen2021transformer} & 63.2 & 57.9 & 67.2 & 67.0 & 51.0 & 55.3 & 65.2 & 56.4 & 63.0 & 58.6 & 62.0 & 64.3 & 64.6 & 62.6 \\
    SeqTrack~\cite{chen2023seqtrack} & 70.9 & 65.9 & 74.6 & 74.9 & 55.1 & 63.0 & 71.4 & 65.2 & 70.5 & 67.2 & 69.7 & 72.4 & 72.4 & 74.9 \\
    EVPTrack~\cite{shi2024explicit} & 71.3 & 64.8 & 75.6 & 73.6 & 61.1 & 65.1 & 72.1 & 66.1 & 71.5 & 67.8 & 70.1 & 72.0 & 72.5 & 74.5 \\
    ODTrack~\cite{zheng2024odtrack} & 71.9 & 66.6 & 76.1 & 73.8 & 60.6 & 65.1 & 72.8 & 67.0 & 70.9 & \textbf{69.3} & 70.7 & 72.2 & 73.0 & \textbf{75.5} \\
    AQATrack~\cite{xie2024autoregressive} & 71.3 & 64.8 & 75.6 & 73.6 & 61.1 & 65.1 & 72.1 & 66.1 & 71.5 & 67.8 & 70.1 & 72.0 & 72.5 & 74.5 \\
    SAM2.1~\cite{ravi2024sam} & 66.3 & 70.1 & 73.0 & 70.5 & 63.6 & 70.9 & 65.5 & 68.6 & 70.6 & 63.0 & 61.3 & 66.3 & 62.9 & 70.0 \\
    \midrule
    % \rowcolor{lightblue}
    SUMO-L & \textbf{73.5} & \textbf{73.2} & \textbf{76.3} & \textbf{75.1} & \textbf{71.7} & \textbf{73.0} & \textbf{73.0} & \textbf{72.1} & \textbf{73.9} & 67.4 & \textbf{72.7} & \textbf{73.8} & \textbf{73.7} & 75.3 \\
    \bottomrule
  \end{tabular}
  }}
  \vspace{-0.1cm}
  \caption{Comparison of attribute-wise AUC(\%) Results for LaSOT~\cite{fan2021lasot}.}
  \label{tab:vot_attribute}
  \vspace{-0.2cm}
\end{table*}

\subsection{Experimental Settings}

\subsubsection{Benchmarks}

We evaluate the \underline{VOT} performance of SUMO on \textbf{LaSOT}~\cite{fan2021lasot}, a large-scale benchmark for long-term tracking; \textbf{LaSOT-ext}~\cite{fan2021lasot}, which extends LaSOT with more challenging and longer sequences; and \textbf{GOT-10k}~\cite{huang2019got}, which focuses on generalization with 10,000+ videos and 560+ object classes.
To evaluate the \underline{MOS} performance of SUMO, we conduct experiments on the established datasets \textbf{SegTrackv2}~\cite{li2013video}, \textbf{FBMS-59}~\cite{ochs2013segmentation}, \textbf{DAVIS2016}~\cite{perazzi2016benchmark}, and \textbf{DAVIS16-MOVING}~\cite{huang2025segment}, which is a subset of DAVIS2016 that focuses on moving objects.

\subsubsection{Metrics}

To evaluate \underline{VOT} performance, we adopt \textbf{AUC}, \textbf{P}$_\textbf{norm}$, \textbf{P}, \textbf{AO}, and \textbf{SR} metrics, as defined in~\cite{huang2019got, muller2018trackingnet}. For \underline{MOS} performance, we use region similarity (\textbf{J}) and contour accuracy (\textbf{F}) metrics, as outlined in~\cite{xie2024appearance}.
Additional implementation details are provided in the supplementary material.

\begin{figure*}[t]
    \centering
    \includegraphics[width=0.95\linewidth]{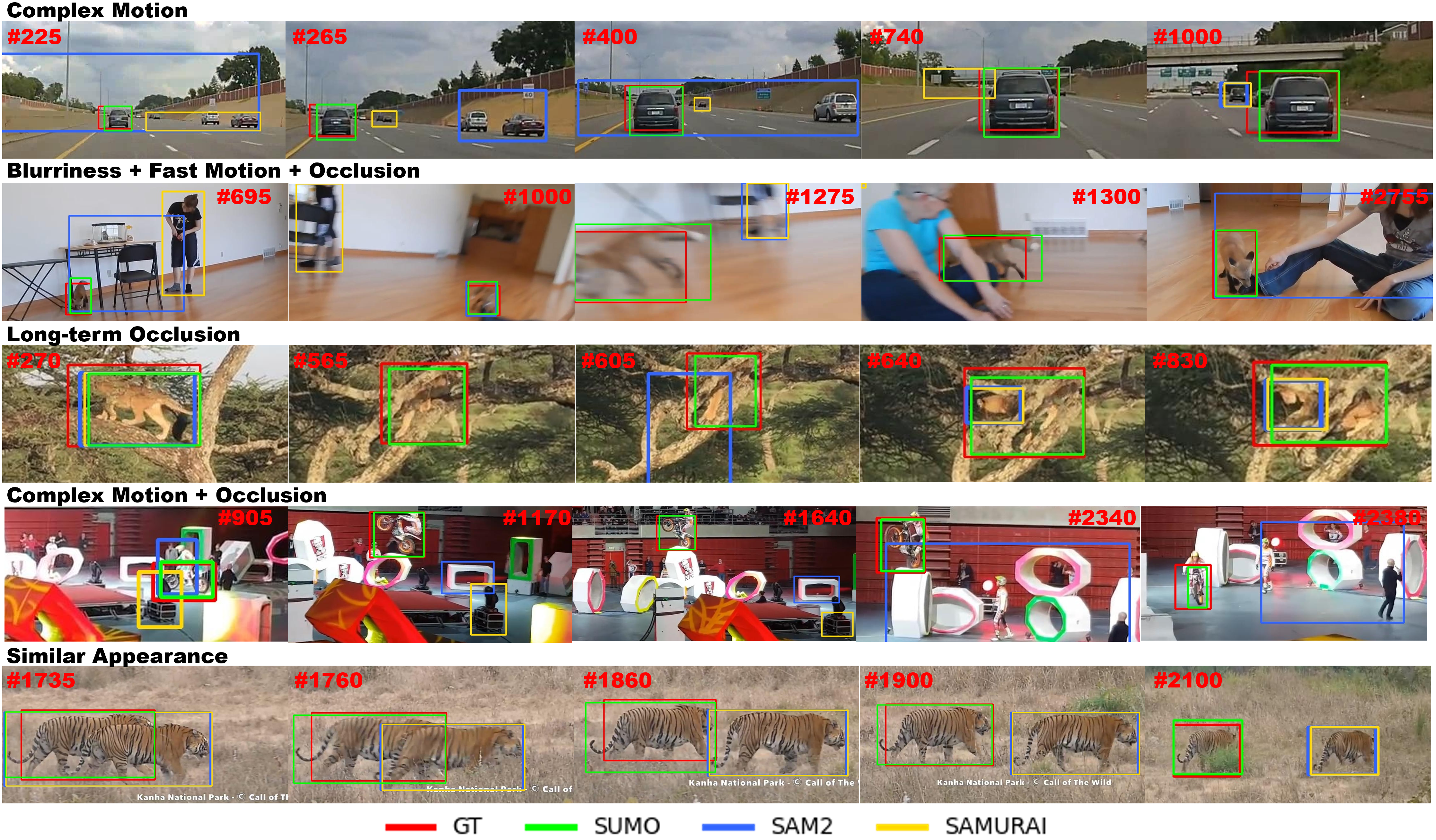}
    \vspace{-0.1cm}
    \caption{Qualitative visualization of VOT performance.}
    \label{fig:qualitative}
    \vspace{-0.2cm}
\end{figure*}

\subsection{VOT Performance}

\subsubsection{Quantitative Performance}

\paragraph{Overall VOT Performance} 

We evaluate the quantitative performance of SUMO on the GOT-10k, LaSOT, and LaSOT-ext benchmarks, comparing it with several strong and widely used baselines, as shown in \cref{tab:tracker_performance}. The results demonstrate that SUMO achieves state-of-the-art performance in VOT across multiple evaluation metrics. Notably, on the LaSOT-ext benchmark, which serves as a challenging stress test with longer sequences and more complex tracking scenarios, SUMO achieves the highest scores across all three metrics, outperforming the strongest baseline by +2.2\% in AUC and +3.8\% in precision. These results further highlight SUMO’s strong performance in long-sequence tracking, its robustness in complex environments, and its ability to generalize effectively to unseen scenarios.

\paragraph{Attribute-wise Performance} 

To further evaluate SUMO’s performance under challenging tracking conditions such as occlusion and deformation, we conduct a detailed attribute-wise analysis on the LaSOT benchmark following the protocol defined in~\cite{fan2021lasot}. These attributes are designed to capture various difficult scenarios commonly encountered in real-world tracking tasks. As shown in \cref{tab:vot_attribute}, SUMO consistently achieves substantial improvements under these complex conditions compared to strong baselines, demonstrating its superior robustness in handling difficult tracking situations. These results further highlight the effectiveness of the proposed Nonlinear SSM and SUF in modeling complex object dynamics and maintaining stable tracking performance under challenging scenarios.

\subsubsection{Qualitative Performance}

\Cref{fig:qualitative} presents qualitative VOT results of SUMO in comparison with SAM2 and SAMURAI across a variety of challenging scenarios, including complex motion, fast motion, long-term occlusion, blurriness, and similar appearance. These examples illustrate typical difficulties encountered in real-world tracking tasks. Overall, SUMO demonstrates robust and accurate VOT performance, consistently maintaining tight alignment with the ground truth throughout the tracking process. It effectively handles diverse challenges by reliably identifying the correct object and precisely localizing it, even under severe visual disturbances. These qualitative results further highlight SUMO’s strong resilience and stability in complex real-world conditions.

\begin{table*}
  \centering{
  \scriptsize
  \resizebox{0.9\linewidth}{!}{
  \begin{tabular}{llccccccccc}
    \specialrule{1.2pt}{0pt}{0pt}
    \toprule
    \multirow{2}{*}{Methods} & \multirow{2}{*}{Source} 
    & \makecell{SegTrackv2} 
    & \multicolumn{2}{c}{FBMS-59} 
    & \multicolumn{3}{c}{DAVIS16-MOVING} 
    & \multicolumn{3}{c}{DAVIS2016} \\
    
    \cmidrule(lr){3-3} \cmidrule(lr){4-5} \cmidrule(lr){6-8} \cmidrule(lr){9-11}
    & 
    & \makecell{$\mathcal{J}\uparrow$} 
    & \makecell{$\mathcal{J}\uparrow$} & \makecell{$\mathcal{F}\uparrow$} 
    & \makecell{$\mathcal{J}\&\mathcal{F}\!\uparrow$} & \makecell{$\mathcal{J}\uparrow$} & \makecell{$\mathcal{F}\uparrow$} 
    & \makecell{$\mathcal{J}\&\mathcal{F}\!\uparrow$} & \makecell{$\mathcal{J}\uparrow$} & \makecell{$\mathcal{F}\uparrow$} \\
    \midrule
    CIS~\cite{yang2019unsupervised} & CVPR'19 & 62.0 & 63.6 & -    & 66.2 & 67.6 & 64.8 & 68.6 & 70.3 & 66.8 \\
    OCLR-flow~\cite{xie2022segmenting} & NIPS'22 & 67.6 & 65.5 & 64.9 & 70.0 & 70.0 & 70.0 & 71.2 & 72.0 & 70.4 \\
    OCLR-TTA~\cite{xie2022segmenting} & NIPS'22 & 72.3 & 69.9 & 68.3 & 78.5 & 80.2 & 76.9 & 78.8 & 80.8 & 76.8 \\
    EM~\cite{meunier2022driven} & PAMI'23 & 55.5 & 57.9 & 56.0 & 75.2 & 76.2 & 74.3 & 70.0 & 69.3 & 70.7 \\
    RCF-Stage1~\cite{lian2023bootstrapping} & CVPR'23 & 76.7 & 69.9 & -    & 77.3 & 78.6 & 76.0 & 78.5 & 80.2 & 76.9 \\
    RCF-All~\cite{lian2023bootstrapping} & CVPR'23 & 79.6 & 72.4 & -    & 79.6 & 81.0 & 78.3 & 80.7 & 82.1 & 79.2 \\
    ABR~\cite{xie2024appearance} & ECCV'24 & 76.6 & 81.9 & 79.6 & 72.0 & 70.2 & 73.7 & 72.5 & 71.8 & 73.2 \\
    Huang et al.~\cite{huang2025segment} & CVPR'25 & 76.3 & 78.3 & 82.8 & 89.5 & 89.2 & 89.7 & 90.9 & 90.6 & 91.0 \\
    \midrule
    SUMO-S & Ours & 84.1 & 89.0 & 89.4 & 85.0 & 83.2 & 86.8 & 87.3 & 86.0 & 88.6 \\
    SUMO-B & Ours & 86.7 & 89.5 & 90.5 & 87.8 & 84.7 & 90.8 & 89.5 & 87.1 & 91.9 \\
    SUMO-L & Ours & \textbf{88.4} & \textbf{90.1} & \textbf{91.3} & \textbf{90.5} & \textbf{89.9} & \textbf{91.0} & \textbf{91.3} & \textbf{90.7} & \textbf{92.0} \\
    \bottomrule
  \end{tabular}
  }}
  \caption{Comparison of MOS performance across SegTrackv2~\cite{li2013video}, FBMS-59~\cite{ochs2013segmentation}, DAVIS16-MOVING~\cite{huang2025segment}, and DAVIS2016~\cite{perazzi2016benchmark} datasets.}
  \label{tab:mos_performance}
\end{table*}

\subsection{MOS Performance}

\subsubsection{Quantitative Performance}

\Cref{tab:mos_performance} illustrates the quantitative MOS performance of SUMO in comparison with several strong baselines on the SegTrackv2, FBMS-59, DAVIS16-MOVING, and DAVIS2016 datasets. These benchmarks cover a variety of motion segmentation scenarios and provide a comprehensive evaluation of segmentation accuracy and temporal consistency. Notably, SUMO achieves state-of-the-art performance in MOS, consistently outperforming prior methods across all datasets in both region similarity ($\mathcal{J}$) and contour accuracy ($\mathcal{F}$). Specifically, SUMO surpasses the previous best by +8.8\% on SegTrackv2 ($\mathcal{J}$), +8.5\% on FBMS-59 ($\mathcal{F}$), and +1.0\% on DAVIS2016-MOVING ($\mathcal{J\&F}$). These results clearly demonstrate SUMO’s effectiveness in achieving temporally consistent and accurate motion segmentation, highlighting its strong generalization capability across diverse and challenging scenarios.

\subsubsection{Qualitative Performance}

\begin{figure}[h]
    \centering
    \includegraphics[width=0.85\columnwidth]{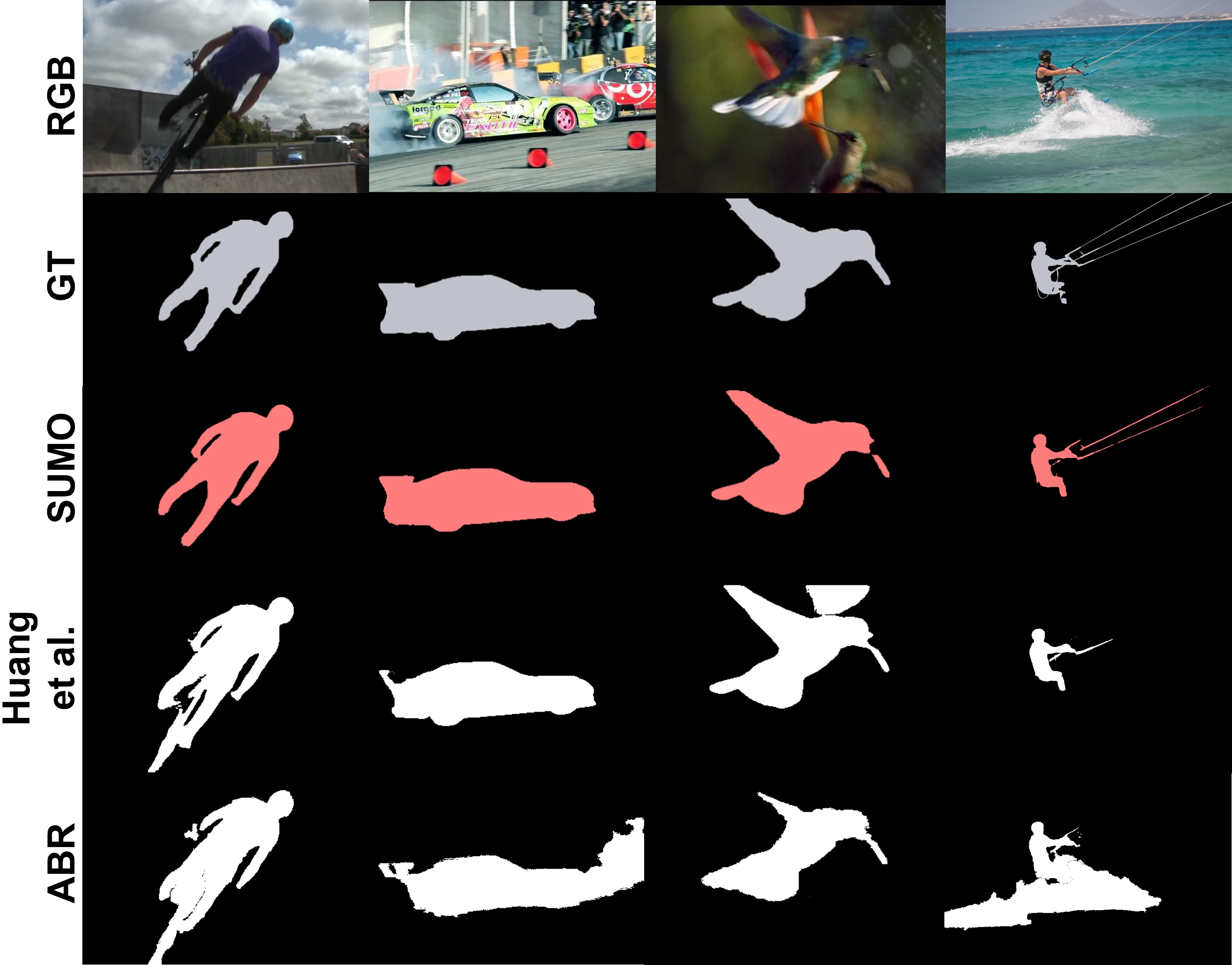}
    \vspace{-0.1cm}
    \caption{Qualitative visualization of MOS performance.}
    \label{fig:qual_mos}
    \vspace{-0.4cm}
\end{figure}

The qualitative results of MOS performance are presented in \cref{fig:qual_mos}. Compared to the baselines, SUMO generates precise segmentation masks with accurate object contours, clearly separating the target object from the background and nearby distractors. In particular, SUMO preserves fine-grained object boundaries while maintaining stable segmentation across frames. This capability is enabled by the inclusion of object shape in the state vector of the nonlinear SSM, as defined in Eq.~\eqref{eq:state_vector}, which jointly models both object shape and temporal dynamics during the tracking process. As a result, the model can effectively capture the evolving geometry of moving objects while maintaining temporal consistency. Overall, SUMO demonstrates accurate, shape-aware, and motion-consistent segmentation of moving objects across diverse challenging scenarios.

\subsection{Ablation Studies}

\subsubsection{Effect of Core Components on VOT}

As shown in \cref{tab:ablation_vot}, we ablate three key SUMO components: the nonlinear SSM, SUF, and memory selection mechanism. Replacing the nonlinear SSM with a linear variant while retaining SUF and memory (second row) reduces AUC by 2.8\%, demonstrating the importance of nonlinear dynamics. Removing all three components (first row) causes a 3.5\% drop, confirming their combined benefit. Adding memory selection (third vs. fourth row) improves AUC by 0.4\%, indicating its complementary role in temporal coherence.

\subsubsection{Effect of Core Components on MOS}
% We ablate the nonlinear SSM, SUF, and memory selection mechanism in MOS. Replacing the nonlinear SSM with a linear variant (second row) reduces $\mathcal{J} \&\mathcal{F}$ by 3.2\%, demonstrating the importance of nonlinear dynamics. Removing all components (first row) causes a 3.3\% drop, confirming the benefit of the full model. Disabling memory (third row) yields a smaller 0.2\% decline, indicating its complementary role in motion segmentation. Overall, the nonlinear SSM is crucial for capturing complex object motion and improving MOS performance.

We perform an ablation study on the nonlinear SSM, SUF, and memory selection mechanism in MOS. Replacing the nonlinear SSM with a linear variant (second row) reduces $\mathcal{J} \&\mathcal{F}$ by 3.2\%, highlighting the importance of nonlinear dynamics modeling. Removing all components (first row) causes a 3.3\% drop, confirming the overall benefit of the full model. Disabling memory selection (third row) results in a smaller 0.2\% decline, indicating its complementary role in motion segmentation. Overall, the nonlinear SSM provides the main performance gain, while SUF and memory selection offer additional improvements.

\begin{table}[t]
\centering
\caption{Effect of core components on VOT. Results are reported on LaSOT.}
\vspace{-0.3cm}
\resizebox{\linewidth}{!}{
\begin{tabular}{ccc|ccc}
\toprule
SSM & SUF & Memory & AUC(\%) & $P_{norm}$(\%) & $P$(\%) \\
\midrule
\ding{55} & \ding{55} & \ding{55} & 70.9 \textcolor{red}{(-3.5\%)} & 78.9 \textcolor{red}{(-3.9\%)} & 76.4 \textcolor{red}{(-4.1\%)} \\
Linear & \ding{51} & \ding{51} & 71.6 \textcolor{red}{(-2.8\%)} & 79.4 \textcolor{red}{(-3.4\%)} & 76.9 \textcolor{red}{(-3.6\%)}\\
Nonlinear & \ding{51} & \ding{55} & 74.0 \textcolor{red}{(-0.4\%)} & 82.3 \textcolor{red}{(-0.5\%)} & 80.0 \textcolor{red}{(-0.5\%)}\\
\midrule
Nonlinear & \ding{51} & \ding{51} & \textbf{74.4} & \textbf{82.8} & \textbf{80.5} \\
\bottomrule
\end{tabular}
}
\label{tab:ablation_vot}
\vspace{-0.2cm}
\end{table}

\begin{table}[t]
\centering
\caption{Effect of core components on MOS. Results are reported on DAVIS16-MOVING.}
\vspace{-0.3cm}
\resizebox{\linewidth}{!}{
\begin{tabular}{ccc|ccc}
\toprule
SSM & SUF & Memory & $\mathcal{J}\&\mathcal{F}\!\uparrow$ & $\mathcal{J}\!\uparrow$ & $\mathcal{F}\!\uparrow$ \\
\midrule
\ding{55} & \ding{55} & \ding{55} & 87.2 \textcolor{red}{(-3.3\%)}& 85.1 \textcolor{red}{(-4.8\%)} & 89.2 \textcolor{red}{(-0.8\%)}\\
% \ding{55} & \ding{55} & \ding{51} &  \\
Linear & \ding{51} & \ding{51} & 87.3 \textcolor{red}{(-3.2\%)} & 85.3 \textcolor{red}{(-4.6\%)} & 89.2 \textcolor{red}{(-0.8\%)}\\
Nonlinear & \ding{51} & \ding{55} & 90.3 \textcolor{red}{(-0.2\%)} & 89.7 \textcolor{red}{(-0.2\%)}& 90.6 \textcolor{red}{(-0.4\%)} \\
\midrule
Nonlinear & \ding{51} & \ding{51} & \textbf{90.5} & \textbf{89.9} & \textbf{91.0} \\
\bottomrule
\end{tabular}
}
\label{tab:ablation_mos}
\vspace{-0.5cm}
\end{table}
\section{Conclusions}
\label{sec:conclusions}

In this work, we introduce SUMO, a unified, zero-shot, training-free framework that combines nonlinear dynamics modeling with vision-based segmentation to achieve accurate and temporally consistent performance on both VOT and MOS tasks. While most existing methods are vision-only, real-world objects often exhibit complex and nonlinear motion patterns. To address this, we introduced a nonlinear SSM to better capture the complex object dynamics. Building on the SSM, we proposed a SUF for accurate state estimation, which incorporates a joint scoring mechanism and adaptively fuses predictions from multiple sources to infer the most plausible object state over time. Extensive experiments demonstrate that SUMO achieves state-of-the-art performance on both tasks. On VOT, SUMO outperforms baselines by up to +3.8\% in Precision, while on MOS, it surpasses the previous best by up to +8.8\% in region similarity ($\mathcal{J}$). Qualitative results further indicate that SUMO performs especially well in challenging scenarios, including occlusion, blurriness, and complex motion, and generates fine-grained, precise masks that accurately capture object contours.
{
    \small
    \bibliographystyle{ieeenat_fullname}
    \bibliography{main}
}

% ================= Supplementary material starts here =================

\clearpage

% Optional: make supplement sections A, B, C, ...
\appendix

% Optional: number supplement figures/tables/equations as S1, S2, ...
\setcounter{figure}{0}
\setcounter{table}{0}
\setcounter{equation}{0}
\renewcommand{\thefigure}{S\arabic{figure}}
\renewcommand{\thetable}{S\arabic{table}}
\renewcommand{\theequation}{S\arabic{equation}}

\section{Additional Implementation Details}

\subsection{Computing Environments}
SUMO is a training-free model, with all inference performed in a zero-shot, per-sequence manner. All experiments are conducted using a single NVIDIA RTX 6000 Ada GPU (48GB) and dual AMD EPYC 9554 CPUs with 512GB RAM. Details of the software environment and dependencies are provided in the code.

\subsection{Model Variants}

We present three variants of SUMO: SUMO-S, SUMO-B, and SUMO-L, which correspond to three sizes of the image encoder backbone: Hiera-S, Hiera-B+, and Hiera-L, respectively. The number of parameters for each image encoder variants is reported in Table~\ref{tab:hiera_para}.

\vspace{-2mm}
\begin{table}[h]
\centering
\caption{Parameter amount of the image encoder backbones used in different SUMO variants.}
\label{tab:hiera_para}
\resizebox{0.6\linewidth}{!}{
\begin{tabular}{l|cc}
\toprule
SUMO Variant & Image Encoder & \#Params of Encoder \\
\midrule
SUMO-S & Hiera-S & 35M \\
SUMO-B & Hiera-B+ & 70M \\
SUMO-L & Hiera-L & 214M \\
\bottomrule
\end{tabular}
}
\end{table}
\vspace{-3mm}

\subsection{Model Parameters}

In our implementation of the Selective Unscented Filter (SUF), we set the diagonal process noise covariance matrix $\Sigma_{rr} \in \mathbb{R}^{n \times n}$, used in the prediction step and defined in Eq.~(19) of the main paper, as:

\begin{equation}
\Sigma_{rr} = 
\begin{bmatrix}
\frac{1}{20} & 0 & 0 & 0 & 0 & 0 & 0 & 0 \\
0 & \frac{1}{20} & 0 & 0 & 0 & 0 & 0 & 0 \\
0 & 0 & 10^{-2} & 0 & 0 & 0 & 0 & 0 \\
0 & 0 & 0 & \frac{1}{20} & 0 & 0 & 0 & 0 \\
0 & 0 & 0 & 0 & \frac{1}{160} & 0 & 0 & 0 \\
0 & 0 & 0 & 0 & 0 & \frac{1}{160} & 0 & 0 \\
0 & 0 & 0 & 0 & 0 & 0 & 10^{-5} & 0 \\
0 & 0 & 0 & 0 & 0 & 0 & 0 & \frac{1}{160}
\end{bmatrix}
\end{equation}

We set the diagonal observation noise covariance matrix $\Sigma_{qq} \in \mathbb{R}^{m \times m}$, used in the update step and detailed in Eq.~(25) of the main paper, as:

\begin{equation}
\Sigma_{qq} =
\begin{bmatrix}
\frac{1}{160} & 0 & 0 & 0 \\
0 & \frac{1}{160} & 0 & 0 \\
0 & 0 & 10^{-5} & 0 \\
0 & 0 & 0 & \frac{1}{160}
\end{bmatrix}
\end{equation}

% time-interval 

%Sigma Point Generation
In the sigma point generation step of SUF, based on the state vector detailed in Eq.~(3) of the main paper, the state dimension is $n = 8$. Consequently, $2n + 1 = 17$ sigma points are generated. To construct the weights $w_i^{(m)}$, $w_i^{(c)}$, and the scaling factor $\lambda$, as detailed in Eqs.~(15)--(16) of the main paper, we set the hyperparameters $\alpha$, $\kappa$, and $\beta$ as:

\begin{equation}
\alpha = 0.1, \quad \beta = 2, \quad \kappa = 0
\end{equation}

For the joint scoring function detailed in Eq.~(22) of the main paper, we set the coefficient $\lambda_{joint}$ as: 
\begin{equation}
\lambda_{joint} = 0.35
\end{equation}

As described in the main paper, after selecting the final mask, we apply a stability threshold $\tau$ on appearance confidence to determine whether to perform the SUF update. In our implementation, we set the threshold $\tau$ as:
\begin{equation}
\tau = 0.3
\end{equation}

In the memory selection mechanism, we set the mask decoder score threshold $\tau_{\text{md}}$, the object confidence threshold $\tau_{\text{obj}}$, and the NSSM score threshold $\tau_{\text{nssm}}$ as follows:

\begin{equation}
\tau_{\text{md}} = 0.5, \quad
\tau_{\text{obj}} = 0.1, \quad
\tau_{\text{nssm}} = 0.5
\end{equation}

% initialization
For the initialization of the SUF, given the observation of the initial frame $\mathbf{z}_0 \in \mathbb{R}^4$, which serves as the initial measurement, we initialize the state mean $\boldsymbol{\mu}_0 \in \mathbb{R}^8$ by concatenating $\mathbf{z}_0$ with a zero velocity vector:
\begin{equation}
\boldsymbol{\mu}_0 =
\begin{bmatrix}
\mathbf{z}_0 \\
\mathbf{0}
\end{bmatrix}
\in \mathbb{R}^8
\end{equation}

The initial covariance matrix $\Sigma_0 \in \mathbb{R}^{8 \times 8}$ is set as a diagonal matrix with the following standard deviations:
\begin{equation}
\Sigma_0 =
\begin{bmatrix}
\frac{1}{10} & 0 & 0 & 0 & 0 & 0 & 0 & 0 \\
0 & \frac{1}{10} & 0 & 0 & 0 & 0 & 0 & 0 \\
0 & 0 &  10^{-2} & 0 & 0 & 0 & 0 & 0 \\
0 & 0 & 0 & \frac{1}{10} & 0 & 0 & 0 & 0 \\
0 & 0 & 0 & 0 & \frac{1}{160} & 0 & 0 & 0 \\
0 & 0 & 0 & 0 & 0 & \frac{1}{160} & 0 & 0 \\
0 & 0 & 0 & 0 & 0 & 0 &  10^{-5} & 0 \\
0 & 0 & 0 & 0 & 0 & 0 & 0 & \frac{1}{160}
\end{bmatrix}
\end{equation}

\subsection{Input Settings}

Our method uses the bounding box of the target object in the first frame as input prompt. The bounding box is represented in the format $[x, y, w, h]$, where $(x, y)$ denotes the center coordinate of the box and $(w, h)$ denotes its width and height, respectively. This initialization is performed once at the beginning of each video sequence and remains fixed throughout inference.

% \section{Evaluation Protocol}

% number of runs

% statistical analysis beyong mean/median

\section{Additional Benchmarks and Metrics Details}

\subsection{VOT Benchmark Details}

\paragraph{LaSOT}is a large-scale benchmark with 1,400 videos densely annotated at every frame. It covers diverse object categories and tracking challenges such as occlusion, scale variation, and background clutter. We evaluate on the official test set of 280 sequences.

\paragraph{LaSOT-ext}extends LaSOT with 150 longer and more difficult sequences. These videos exhibit more severe occlusions, abrupt motion, and low-resolution targets. We use the full set of 150 sequences for evaluation.

\paragraph{GOT-10k}contains over 10,000 videos and 1.5 million bounding boxes across 560+ object categories. It adopts a one-shot protocol where test classes are disjoint from training, emphasizing generalization. We perform inference on the official test set of 180 videos.

\subsection{MOS Benchmarks Details}

\paragraph{SegTrackv2}includes 14 short videos with pixel-level annotations of deformable, fast-moving objects. The dataset features low resolution, non-rigid motion, and challenging boundaries. We evaluate on the full test set.

\paragraph{FBMS-59}consists of 59 videos with sparse annotations, focusing on motion-consistent grouping rather than appearance. It emphasizes long-term temporal consistency. We use the 30-video test split for evaluation.

\paragraph{DAVIS2016}provides 50 high-resolution videos with dense annotations for foreground-background segmentation. It includes deformable objects, fast motion, and complex scenes. We perform inference on the 20-video validation set, following common protocol.

\paragraph{DAVIS16-MOVING}is a subset of DAVIS2016 containing 12 sequences where the main object exhibits significant motion. It is designed to isolate motion-centric segmentation performance. We evaluate on all 12 sequences.

\section{Additional Experiments}

\subsection{Comparison with SAM2}

We compare SUMO with SAM2 to evaluate the effectiveness of integrating our NSSM methods into the tracking and segmentation pipeline. Experiments are conducted on both the LaSOT benchmark (for VOT) and the DAVIS16-MOVING benchmark (for MOS). Results are summarized in Table~\ref{tab:sam2_comparison}. Across all model sizes (Small, Base, and Large), SUMO consistently outperforms SAM2 on both benchmarks. On LaSOT, SUMO achieves clear improvements in AUC, normalized precision ($P_\text{norm}$), and precision ($P$), indicating better accuracy and stability in long-term object tracking. On DAVIS16-MOVING, SUMO also shows notable gains in both region similarity ($\mathcal{J}$) and contour accuracy ($\mathcal{F}$), as well as the combined $\mathcal{J}\&\mathcal{F}$ score. These results validate that the integration of nonlinear dynamics into the vision framework significantly improves performance on both VOT and MOS tasks.

\begin{table}[t]
  \centering
  \resizebox{\linewidth}{!}{
  \begin{tabular}{l|cccccc}
    \specialrule{1.2pt}{0pt}{0pt}
    \toprule
    \multirow{2}{*}{Methods} 
    & \multicolumn{3}{c}{LaSOT (VOT)} 
    & \multicolumn{3}{c}{DAVIS16-MOVING (MOS)} \\    
    \cmidrule(lr){2-4} \cmidrule(lr){5-7}
    & \makecell{$AUC\uparrow$} & \makecell{$P_{norm}\uparrow$} & \makecell{$P\uparrow$} 
    & \makecell{$\mathcal{J}\&\mathcal{F}\!\uparrow$} & \makecell{$\mathcal{J}\uparrow$} & \makecell{$\mathcal{F}\uparrow$} \\
    \midrule
    SAM2-S & 66.9 & 56.1 & 83.4 & 84.4 & 82.8 & 86.1\\
    SUMO-S & 70.9 & 78.3 & 76.1 & 85.0 & 83.2 & 86.8\\
    \midrule
    SAM2-B & 67.1 & 55.3 & 83.0 & 87.0 & 83.5 & 89.5 \\
    SUMO-B & 72.8 & 81.0 & 78.6 & 87.8 & 84.7 & 90.8\\
    \midrule
    SAM2-L & 71.0 & 53.7 & 85.9 & 86.9 & 85.0 & 88.3 \\
    SUMO-L & 74.4 & 82.8 & 80.5 & 90.5 & 89.9 & 91.0\\
    \midrule
    \bottomrule
  \end{tabular}
  }
    \caption{Comparison between SUMO and SAM2 across different model sizes (Small, Base, Large) on LaSOT (VOT) and DAVIS16-MOVING (MOS).}

  \label{tab:sam2_comparison}
\end{table}
\vspace{0mm}

\section{Additional VOT and MOS Visualizations}

\subsection{VOT Visualization}

\begin{figure*}[b]
    \centering
    \includegraphics[width=\linewidth]{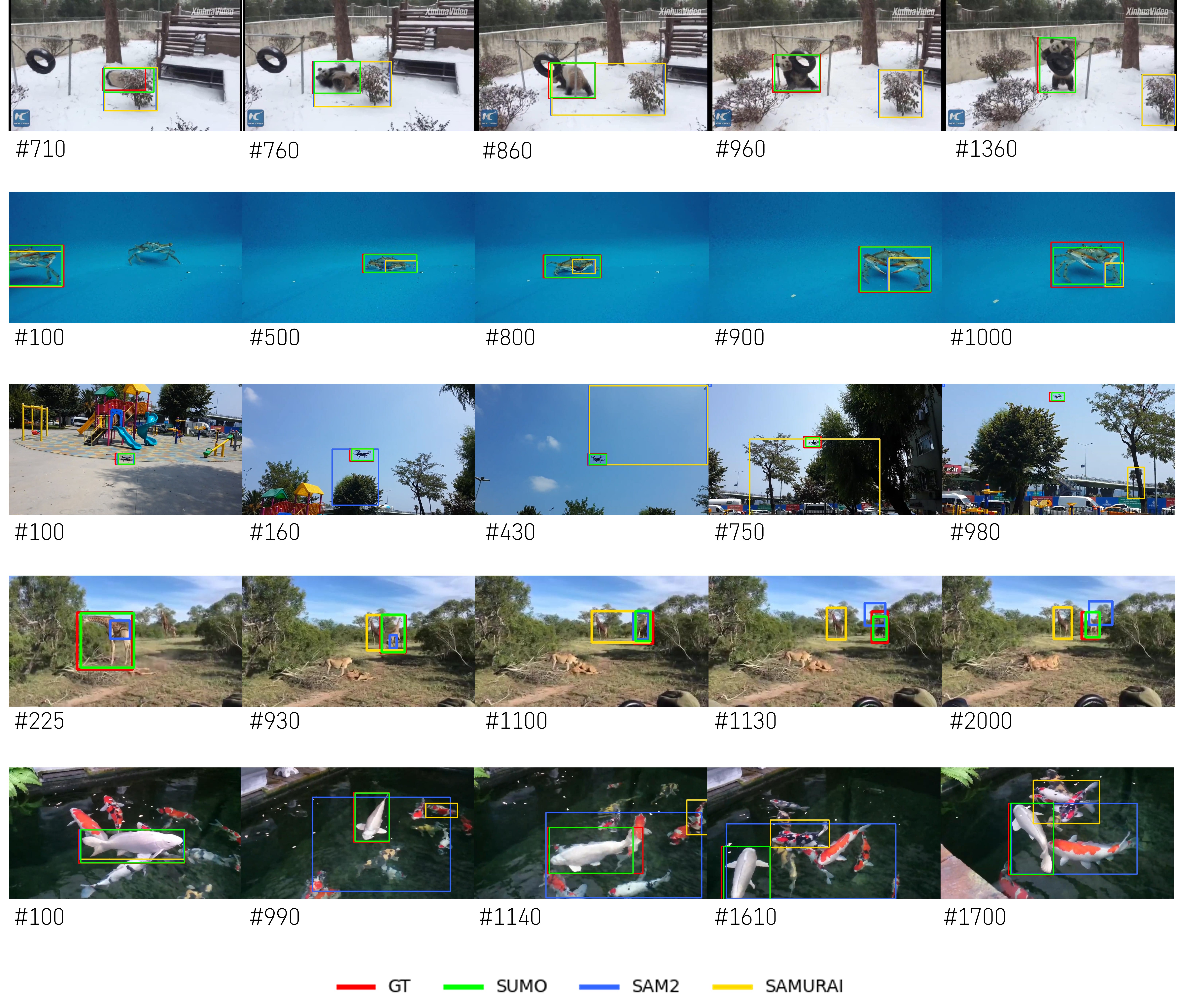}
    \caption{Qualitative visualization of VOT performance.}
    \label{fig:vot1}
\end{figure*}

\begin{figure*}
    \centering
    \includegraphics[width=\linewidth]{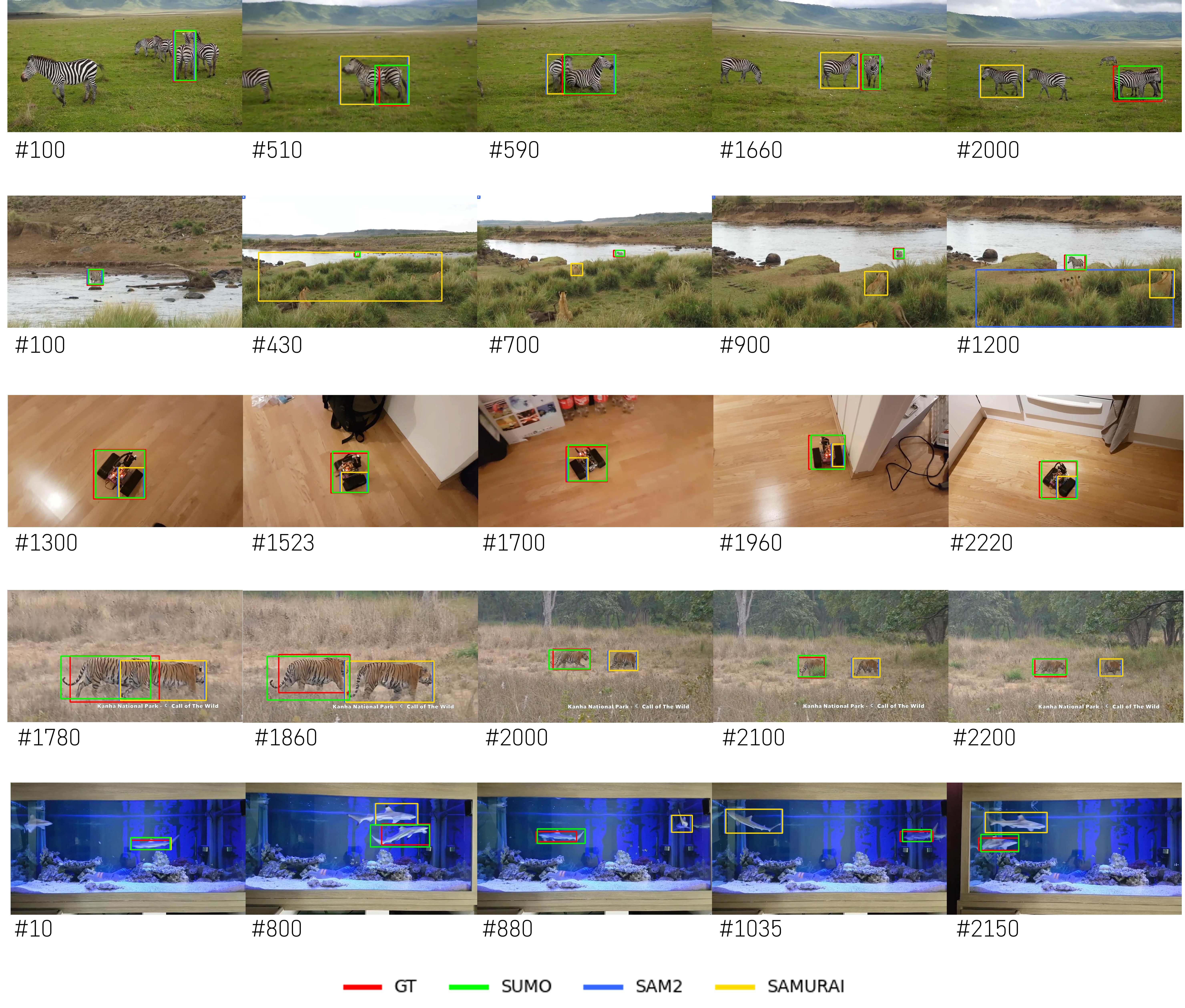}
    \caption{Qualitative visualization of VOT performance.}
    \label{fig:vot2}
\end{figure*}

Figure~\ref{fig:vot1} and Figure~\ref{fig:vot2} present additional VOT visualizations covering diverse and challenging scenarios. SUMO exhibits strong performance under complex motion, occlusion, and shape deformations in long-term tracking, consistently achieving superior localization accuracy and robustness compared to the baselines.

\section{MOS visualization}

\begin{figure*}
    \centering
    \includegraphics[width=\linewidth]{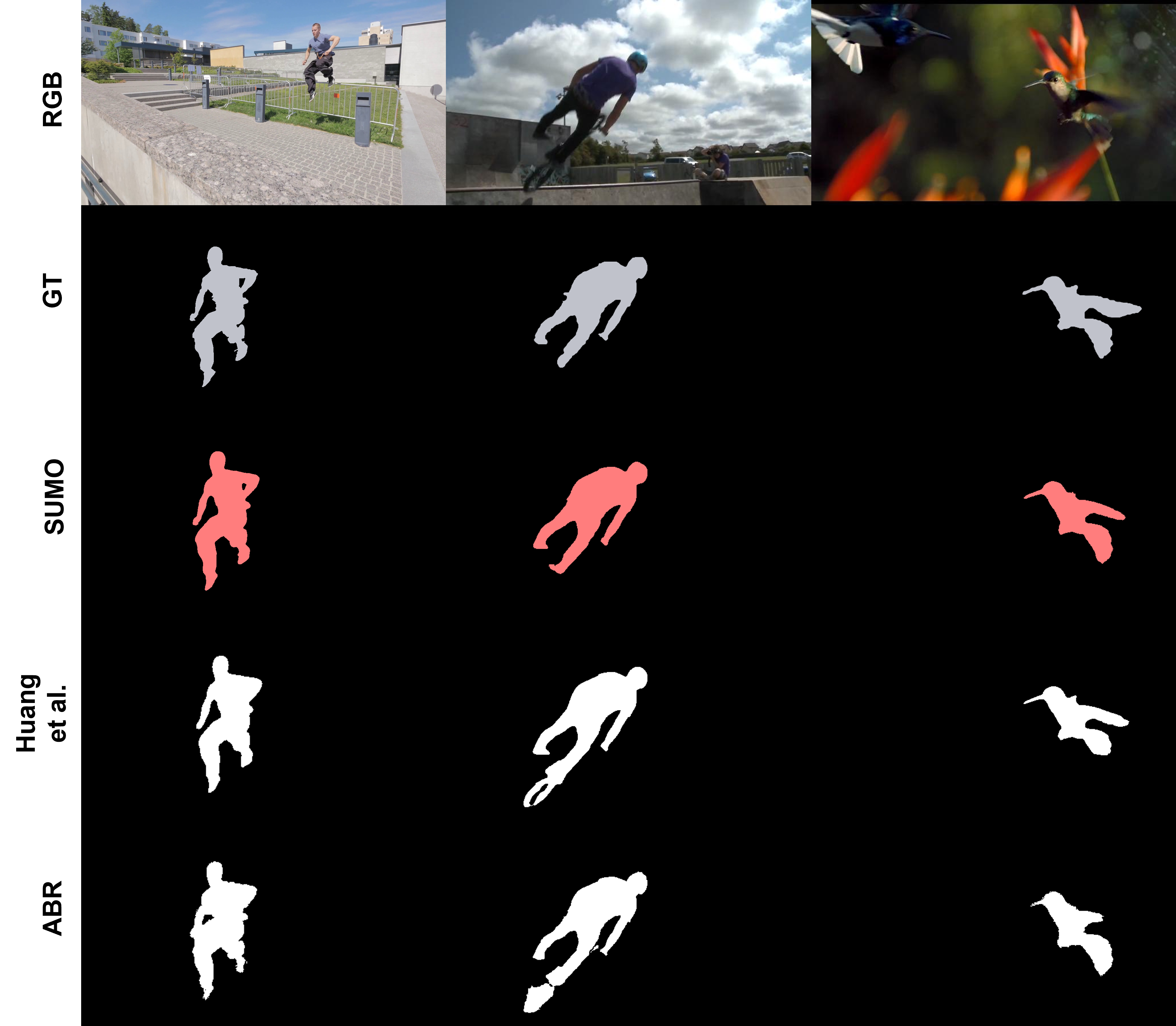}
    \caption{Qualitative visualization of MOS performance.}
    \label{fig:mos1}
\end{figure*}

\begin{figure*}
    \centering
    \includegraphics[width=\linewidth]{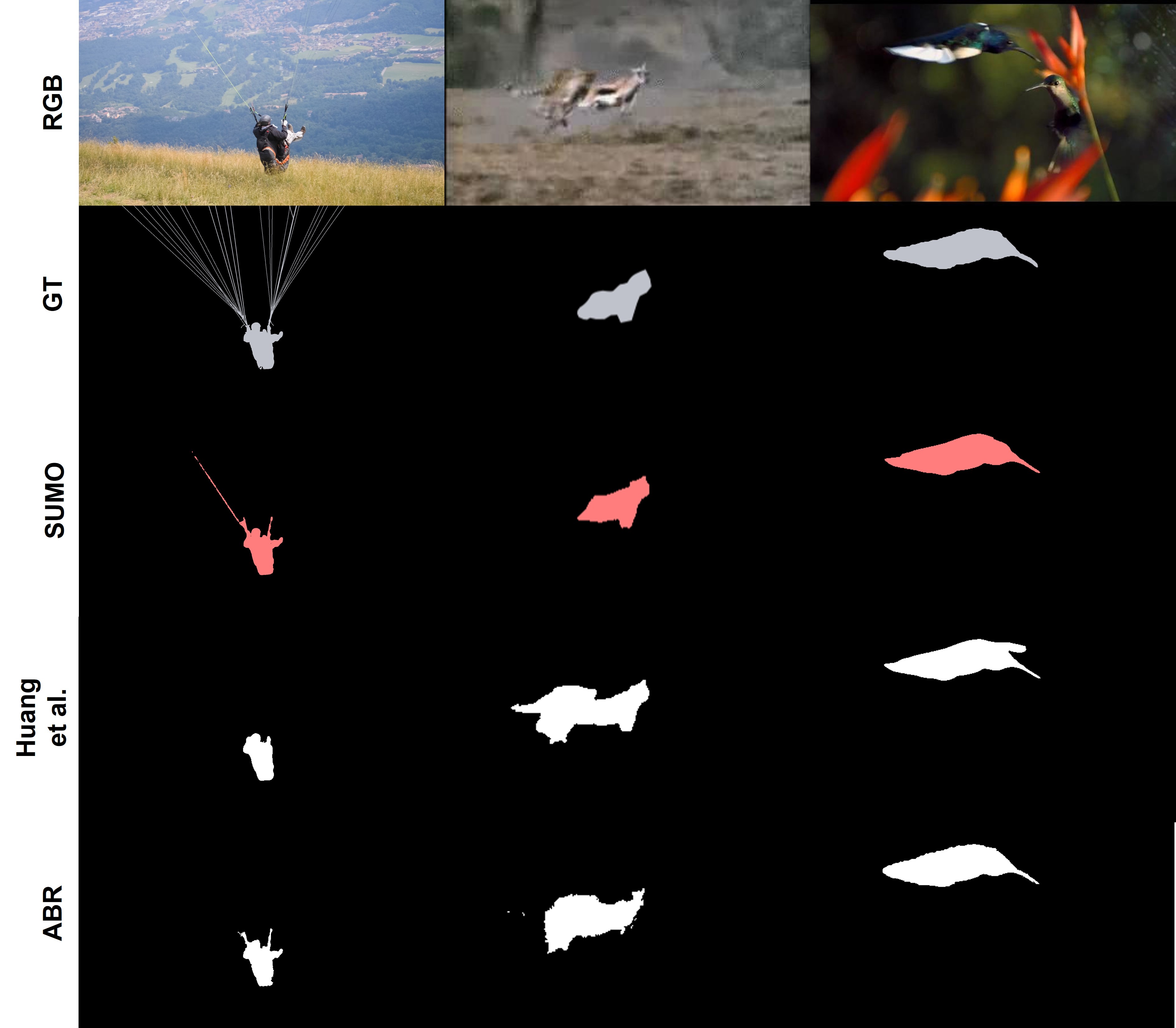}
    \caption{Qualitative visualization of MOS performance.}
    \label{fig:mos2}
\end{figure*}

\begin{figure*}
    \centering
    \includegraphics[width=\linewidth]{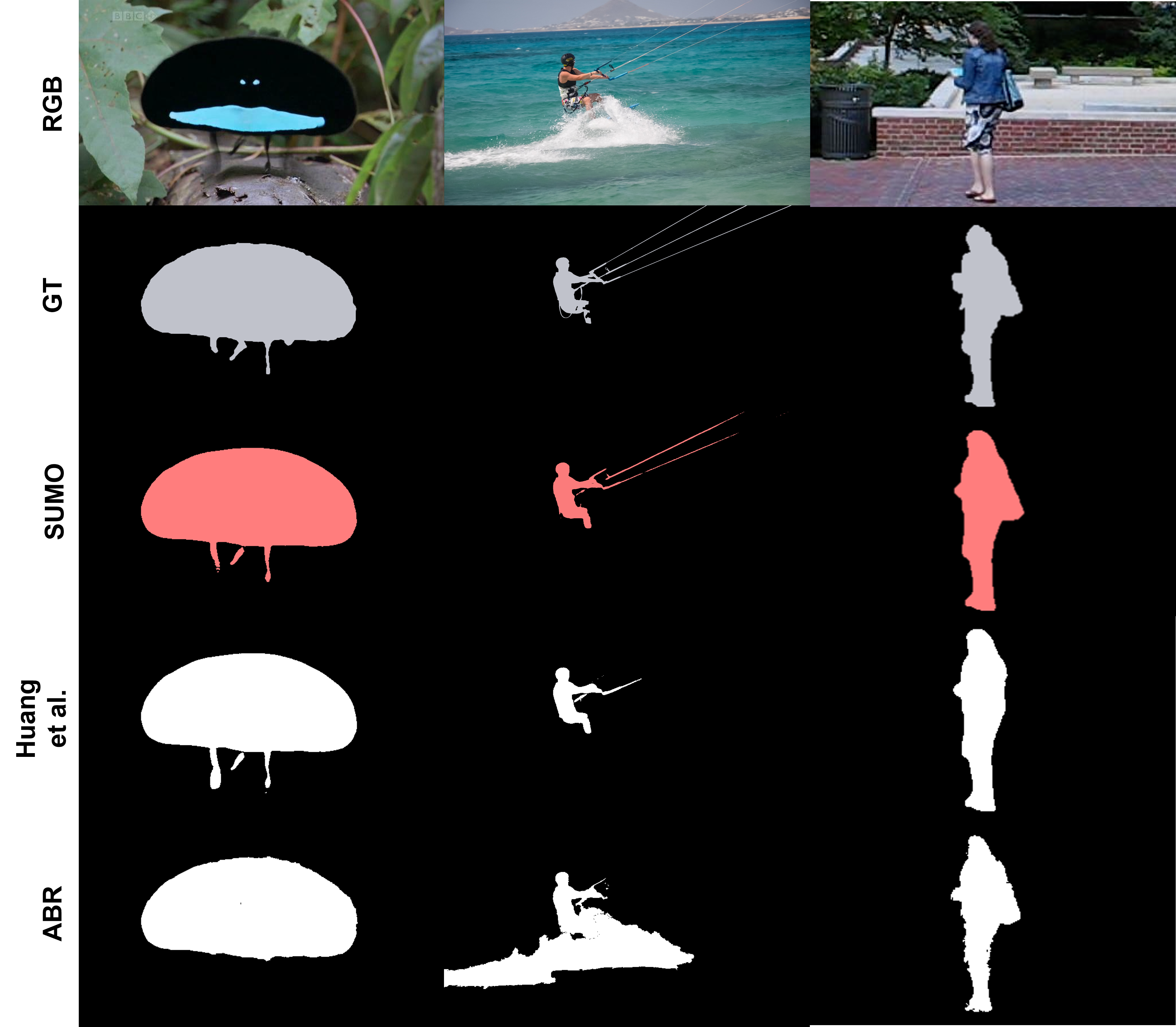}
    \caption{Qualitative visualization of MOS performance.}
    \label{fig:mos3}
\end{figure*}

Figure~\ref{fig:mos1}, Figure~\ref{fig:mos2}, and Figure~\ref{fig:mos3} present additional MOS visualizations across various challenging scenes. SUMO produces high-quality segmentation masks with precise contours, accurate object boundaries, and minimal inclusion of irrelevant regions. It consistently segments the correct target object with high spatial precision, outperforming baselines in terms of mask fidelity and object completeness.

\clearpage

\end{document}